\documentclass[letterpaper, 10 pt, conference]{ieeeconf}  % Comment this line out if you need a4paper

\IEEEoverridecommandlockouts                              % This command is only needed if 
\usepackage{graphicx}
\usepackage{multirow}
\usepackage{ascmac}
\usepackage{amsmath}
\usepackage{url}

\title{\LARGE \bf
What Should the System Do Next?: \\
Operative Action Captioning for Estimating System Actions
}

\author{Taiki Nakamura$^{1,2}$, Seiya Kawano$^{1}$, Akishige Yuguchi$^{1}$, Yasutomo Kawanishi$^{1}$, and Koichiro Yoshino$^{1}$% <-this % stops a space
\thanks{*This work was not supported by any organization}% <-this % stops a space
\thanks{$^{1}$Guardian Robot Project, R-IH, RIKEN, 2-2-2, Hikaridai, Seika, Sohraku, Kyoto, 6190288, Japan
        {\tt\small \{seiya.kawano, akishige.yuguchi, yasutomo.kawanishi, koichiro.yoshino\} @riken.jp}}%
\thanks{$^{2}$Graduate School of Information Science and Technology, The University of Tokyo, 7-3-1 Hongo, Bunkyo-ku, Tokyo 113-8656, Japan
        {\tt\small supikiti@g.ecc.u-tokyo.ac.jp}}%
}

\begin{document}

\maketitle
\thispagestyle{empty}
\pagestyle{empty}

%%%%%%%%%%%%%%%%%%%%%%%%%%%%%%%%%%%%%%%%%%%%%%%%%%%%%%%%%%%%%%%%%%%%%%%%%%%%%%%%
\begin{abstract}

%ロボットをはじめとする人間を支援するシステムは、観測から状況を正しく理解し、人間が必要とする支援行動を出力する必要がある。
%特に人間を対象とした支援において、システムがどのような状況理解を行い、どのような動作行動の生成しようとしているかは、言語で表現することが重要である。
%そこで本研究では、現在の状況からシステムが行うべき行動を推定しその内容を言語で説明する、動作行動推定とその言語化 (captioning operative action) に取り組む。
%具体的には、ある状況とそこに対して何らかの支援行動が行われた理想状態の画像を入力とし、どのような支援行動が行われたかを説明する言語化タスクによって動作行動推定を実現するシステムを構築した。
%この際、こうした状況を説明する補助情報であるシーングラフの推定を補助タスクとして用いることで、シーングラフのアノテーションが存在しないテストセットに対しても精度高く動作行動の推定・言語化を行うことができることが確認された。

Such human-assisting systems as robots need to correctly understand the surrounding situation based on observations and output the required support actions for humans.
%Expressing the internal states of robots, which include understanding and action planning results, by natural language is essential because it is difficult for humans to understand or observe a robot's internal states. 
Language is one of the important channels to communicate with humans, and the robots are required to have the ability to express their understanding and action planning results.
In this study, we propose a new task of operative action captioning that estimates and verbalizes the actions to be taken by the system in a human-assisting domain. 
We constructed a system that outputs a verbal description of a possible operative action that changes the current state to the given target state. 
We collected a dataset consisting of two images as observations, which express the current state and the state changed by actions, and a caption that describes the actions that change the current state to the target state, by crowdsourcing in daily life situations.
Then we constructed a system that estimates operative action by a caption. 
Since the operative action's caption is expected to contain some state-changing actions, we use scene-graph prediction as an auxiliary task because the events written in the scene graphs correspond to the state changes. 
Experimental results showed that our system successfully described the operative actions that should be conducted between the current and target states.
The auxiliary tasks that predict the scene graphs improved the quality of the estimation results.

%By using the estimation of scene graphs, which are auxiliary information to explain such situations, as an auxiliary task, it was confirmed that the system can estimate and verbalize behavioral behavior with high accuracy even for test sets without scene graph annotations.

\end{abstract}

%%%%%%%%%%%%%%%%%%%%%%%%%%%%%%%%%%%%%%%%%%%%%%%%%%%%%%%%%%%%%%%%%%%%%%%%%%%%%%%%
\section{Introduction}
\label{sec:intro}
%近年の深層学習技術の進展から、画像、信号、その他様々な観測を言語によって理解するキャプショニングの研究が進められている\cite{vinyals2016show,you2016image}。
%言語によって画像や信号に説明を加えることで、機械は人間が理解可能な形でデータへの解釈を与えることができ、様々なアプリケーションへ応用することができる\cite{dou2018data2text,ishigaki2021generating}。

Recent advances in deep learning technology have fueled to research on situation understanding in which images, signals, and various other observations are understood through language~\cite{vinyals2016show,you2016image}.
Systems can provide interpretations of data in a form comprehensible to humans by adding explanations to data through language for a variety of applications~\cite{dou2018data2text,ishigaki2021generating}.
%
%こうしたシステムの応用先は様々であるが、その応用先の一つとして生活支援ロボットなどの生活空間で動作するシステムがある。
%こうしたシステムは人間との共生空間で動作するため、言語という人間にとって解釈可能な形で状況理解を行うことには大きな意味がある。
%このような言語を用いた状況理解は、観測した人の行動\cite{takano2015statistical}や、ロボットに類する自律動作システムの動作\cite{yamada2018paired,yoshino2020caption}、ロボットが行う多様な観測\cite{yuguchi2022butsukusa}など、様々な状況で検討されている。
%
There are various applications for such systems, one of which is systems that operate in human living spaces, such as life-support robots.
Since these systems operate in a symbiotic space with humans, the systems must understand the situation in a form that can be interpreted by humans, such as natural language.
Such language-based situational understanding has been discussed in a variety of situations, including human behavior analysis~\cite{takano2015statistical}, describing robot behaviors~\cite{yamada2018paired,yoshino2020caption}, and robot observations~\cite{yuguchi2022butsukusa}.

%観測から状況を正しく理解し説明できることは、人間と共生するシステム構築の第一歩である。
%しかし実際には、システムはユーザに対して協調的な支援を提供することが期待されている。
%つまり、単なる状況の理解から一歩進んで、どのような行動を期待されているかについても正しく理解する必要がある。
%例えばロボットの研究では、観測から得られる状況を状態（state）とし、これに対応するロボットの行動クラス（action）を正しく識別することが提案されている\cite{soans2020sa}。
%また、現在の状態と理想状態を入力し、この間に行うべきロボットの行動クラスを推定する問題も定義されている\cite{chatila2018toward}。

Correctly understanding and explaining a situation from observations is the first step in building a system that can work in human living spaces.
However, systems are expected to provide cooperative assistance to human users.
In other words, they must recognize both the current situation and the necessary actions (help) for solving it.
For example, robotics research has proposed to accurately identify the expected robot action class given the current robot observation as input~\cite{soans2020sa,ahn2022can}.
Another study defined the problem of estimating the robot's action class that should be performed between the current and target states~\cite{chatila2018toward}.

%ここで、ロボットに類する生活支援システムは多種多様であり、行うべき支援も多岐に渡ることが問題となる。
%あらかじめ定義されたロボットの行動クラスだけでは、柔軟な状況理解を実現することが難しい。
%そこで本研究では、支援システムが行うべき行動を理解・明示するための第一歩として、 キャプショニングを用いた動作行動推定 (captioning operative action) のタスクを提案する。
%本タスクでは、視覚情報で構成される現在の観測と理想状態をシステムへ入力し、その間でシステムが行うべき行動を言語で説明することで、ユーザ支援システムの柔軟な状況理解と、システムとユーザの適切な情報共有を実現する。

One of the most critical issues of these proposals is that there is a wide variety of life support systems that resemble those of robots, and the actions conducted for such support are also diverse.
In other words, achieving a flexible understanding of a situation is challenging with only predefined robot action classes.
Thus, in this study, we propose an operative action captioning task, which describes what action is to be done between the current to the target state by captioning for a better understanding of the surrounding human situation.
In this task, we assume that the current state (observation) and the target state are given by cameras.
%consist of visual information and are used as input to the system.
Then the system generates a caption that explains the expected actions to change the current state to the target state.
Using this method, such human-assisting systems as robots can interact with users for mutual understanding.

%キャプショニングの研究はコンピュータビジョンの分野で盛んに行われており、ある程度の学習データ量が必要である\cite{kim2019image}。
%そこで本研究ではクラウドソーシングを用いて、17,000件程度の動作行動推定のためのデータセットを構築した。
%また既存のキャプショニングの研究では、限られた学習データでキャプショニングの精度を向上させるために画像から認識可能な情報を補助情報として使うことの重要性が示唆されている\cite{gan2017semantic,yao2017incorporating,li2019pointing}。
%本研究では動作行動推定に必要な知識が時系列上の画像間の変化であるという点に着目し、これらの変化に焦点を置いたシーングラフ (Scene Graph)~\cite{li2017scene}の推定を動作行動推定の補助タスクとして用いる。
%シーングラフは画像中の情報が「subject-relationship-object」の形式で記述され、動作行動推定と相性が良いことを期待する。

Situation understanding by captioning has been actively studied in computer vision, which requires a certain amount of training data~\cite{kim2019image}.
Therefore, we used crowdsourcing to construct a dataset of about 17,000 cases for operative action captioning.
Existing research on captioning suggests the importance of using auxiliary information that can be recognized from images to improve the accuracy of captioning with limited training data~\cite{gan2017semantic,yao2017incorporating,li2019pointing}.
We focus on scene graphs~\cite{li2017scene}, which represent events on images as auxiliary information.
The scene graph represents some events in the image, which are strongly related to the actions conducted in the images.
A scene graph describes the situation in detail as a set of triplets: subject-relationship-object.
By taking information from the scene graph for both the current and target states, we can acquire the differences or the conducted events between two states, which is critical information to acquire the action between them.

%In this study, our task focuses on actions conducted between the current and the target states; thus, we use scene graph~\cite{li2017scene}, which represents events on images, as the auxiliary information.
%The scene graph indicates events on images in a triplet form; ``subject-relationship-object.''
%We expect that the information from the scene graph may contain important information for the operative action captioning task.

%the knowledge necessary for estimating behavioral behavior is the changes between images in a time series, and use scene-graph estimation, which focuses on these changes, as an auxiliary task for behavioral behavior estimation.
%Scene graphs describe information in images in the form of "subject-relationship-object" and are expected to be compatible with behavioral behavior estimation.

%キャプショニングは画像中の事物を説明するもので、画像中の物体情報などを認識することが重要である\cite{}。
%これに対して本研究で扱う動作行動推定では、画像から得られる動作情報や、画像間での変化に相当する情報をいかに得るかが重要となる。
%こうした画像間での変化を表現した情報として、本研究では画像の理解のために用いるシーングラフ (Scene Graph) に着目し、シーングラフの推定を動作行動推定の補助タスクとして利用した。

%本論文ではまずキャプショニングを用いた動作行動推定のタスクについて説明する（2章）．
%このタスクを実現するため、 Home Action Genome Dataset をクラウドソーシングによって拡張してデータを準備した（3章）。
%次に、動作行動推定・言語化を行うキャプショニングモデルの構築を行った（4章）。
%キャプショニングのベースラインモデルとしては Dual Dynamic Attention Model (DUDA) を用い、 Home Action Genome Dataset に付与されている動作を表すシーングラフの推定を補助タスクとして用いるモデルを構築した。
%実験では、実際に補助タスクがあるモデル、無いモデルによって生成された動作行動推定結果を自動評価・人手評価によって比較した（5章）。

Our research contribution follows:
\begin{itemize}
    \item We defined a new task, operative action captioning, toward building a robust robot action generation system.
    \item We built a new dataset for the defined task by extending the existing dataset at home, Home Action Genome Dataset.\footnote{We will open the data in camera ready.}
    \item We proposed a strong baseline based on change captioning and an auxiliary task of scene-graph prediction.
\end{itemize}

%The rest of this paper is organized as follows. 
%Section 2 describes operative action captioning tasks.
%Section 3 shows how to construct a dataset by extending the Home Action Genome Dataset by crowdsourcing.
%Section 4 gives the model and implementation details of our proposed operative action captioning model, which is extended from the baseline model.
%We used the dual dynamic attention (DUDA) model as a baseline and extended it to use auxiliary information from scene graphs.
%Experiments in Section 5 show that our proposed model worked better in the operative action captioning task than the baseline in both automatic and human evaluations.
%We conclude the paper and discuss future work in Section 6.

\begin{figure}[t]
    \centering
    \includegraphics[width=\linewidth]{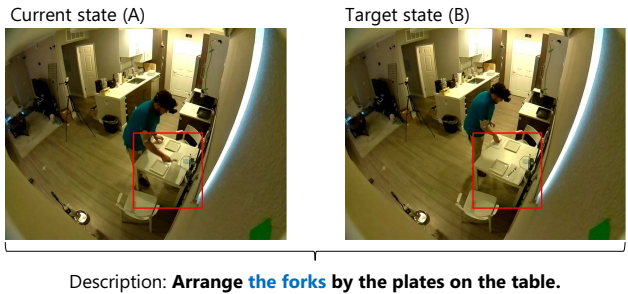}
    \caption{Overview of proposed task: operative action captioning. 
    In the dataset, an assistant worked in a role of human assisting robot has a head-mount camera and images are captured by the camera.
    First-person viewpoint indicates images from the assistant. %, which can be replaced with robot's first-person view camera. Images are captured by head-mount camera of a human-assistant.%; human-assistant systems such as robots can acquire such images.
    Third-person viewpoint indicates images from fixed-point cameras.
    Such information can also be used by systems if cameras are available.
    Red squares indicate rectangles, which contain target objects for the operation.}
    \label{fig:overview}
\end{figure}

\section{Operative Action Captioning}
\label{sec:task}
% タスク設計の概要

%まず、本研究で対象とするタスクの概要を図\ref{fig:overview}に示す。
%システムには支援動作前の状態 (A) と、支援動作後の状態 (B) が与えられる。
%A はロボットなどの生活支援システムが観測している現在の状態であり、 B は支援行動（ここではトイレの便座の蓋を開ける）が終了した時点の理想的な状態である。
%本研究では A および B の画像を入力とし、その間に行われる動作行動を説明文生成により推定する。

%First, an overview of the task targeted in this study is shown in 
Figure~\ref{fig:overview} overviews the operative action captioning task.
%The system is given a state before the support action (A) and a state after the support action (B).
{\sf (A)} is the current state observed by the life support system, such as robots, and {\sf (B)} is the target state when the support action is completed.
In this case, ``opening the toilet seat lid'' is the expected support action, and the system has to distinguish it from both the current and target states.
In this study, images {\sf (A)} and {\sf (B)} are used as input, and the action that changes the state {\sf (A)} to {\sf (B)} is estimated by generating explanatory sentences (caption).

%ロボットの行動クラス推定の問題では、これら2つを入力としてロボットの行うべき行動を推定する\cite{chatila2018toward}。
%つまり、現在の環境 A をどのような行動によって理想状態 B に変更できるかを推定する問題である。
%しかし実際には生活支援で行うべき動作は多岐に渡り、あらかじめ定義されたロボットの行動クラスだけで状況を理解し必要な支援行動を明確化することが難しい場合も存在する。

In a general scenario for robots, the problem is defined as selecting an action class from pre-defined action classes, given these two inputs~\cite {chatila2018toward}.
In other words, the problem is predicting an action class that can change current state {\sf (A)} to target state {\sf (B)}.
However, a wide variety of actions must be performed in daily life support.
This diversity makes it difficult for systems to estimate the support actions that are required for life support systems.
We apply a generative approach based on captioning to estimate the necessary support actions from both pre-defined action classes and undefined actions.

%and there are cases in which it is difficult to understand the situation and clarify the necessary support actions with only the predefined action classes of the robot.

%また、こうした行うべき行動の認識は、観測からの行動認識という文脈でも研究が行われている\cite{tran2018closer}。
%こうした研究の多くでは、画像や動画を入力として、その中に存在する行動のクラスラベルを推定する。
%また、規定の行動クラスだけでなくシーングラフ\cite{li2017scene}など、より複雑な形で動作行動を含む事態を記述し利用するアプローチも存在する\cite{ji2020action,rai2021home}。
%本研究では、より自由度の高い動作行動推定の記述として自然言語による説明文生成を導入する。

Recognition of such operative actions has been researched in the field of possible action recognition from observations~\cite{tran2018closer,soans2020sa,ahn2022can}.
Such studies define the problem by estimating the action class contained in images or videos.
More complicated event representation, such as scene graphs~\cite{li2017scene}, is used to predict actions or events in observations~\cite{ji2020action,rai2021home}.
In this study, we introduce captioning using natural language as a more flexible estimation of operative actions.
%
%また、シーングラフを導入することでキャプショニングの精度が向上することは言われている\cite{chen2020say}。
%そこでは本研究ではこうしたアイディアを取り込み、シーングラフの推定を補助タスクに用いることで、キャプショニングによる動作行動推定の精度を向上させる。
The use of scene graphs also generates accurate and clear captions~\cite{chen2020say}.
Our study is inspired by such works and uses scene-graph prediction as an auxiliary task to improve the generated caption.
Another advantage of using auxiliary learning compared with input representation integration is that we do not need to prepare labels in the test phase.

%異なる環境の画像2種類をキャプショニングに用いる研究では、その多くは画像の差分に着目している\cite{park2019robust,qiu20203d}。
%これに対して本研究では、2つの状態の差から、その状態を変化させる行動のキャプショニングを行う。
%こうした状態を変化させる行動の推定では、ロボットの主体的動作の推定が取り組まれている\cite{kim2021fixmypose}。
%本研究ではロボットの身体性にとらわれず、生活空間におけるより一般的な動作行動の推定を行う。

Captioning differences is another related research direction that focuses on the different descriptions of two images~\cite{park2019robust,qiu20203d}.
In contrast, in this study, our task captions the actions that should be conducted between two states (images).
The difference information is useful to predict the operative actions; thus, we use a captioning system of difference as our baseline.

Raw-level robot action estimation from two such observations is another research avenue~\cite{kim2021fixmypose}.
A robot's physicality strongly constrains such an approach.
Our captioning approach can express the expected action by language, even if the target robot cannot conduct the generated operative action.
This point is critical for robots working in human living space.

%In the estimation of actions that change such states, the estimation of the robot's subjective actions is addressed.
%This study is not restricted to the physicality of the robot, but estimates more general action behaviors in the living space

%\subsection{ロボットの動作行動推定}
%% ロボットドメインでの前後のイメージから何をやるか決めるタスク

%\subsection{動作行動推定と状況理解}
%% 言語での説明が重要

%\subsection{動作行動とシーングラフ}
%% シーングラフを用いた行動推定

\section{Data Collection}
\label{sec:data}
% 集めたデータについて

%今回説明文生成を用いた動作行動推定を実現するため、動作行動に対応する説明文を収集した。
%今回は家庭内ロボットが行うべき行動を説明することを想定するため、 Home Action Genome Dataset\cite{rai2021home} を拡張して用いた。
%本節ではまず Home Action Genome Dataset について説明する。
%次に、データセットに対してクラウドソーシングを用いて動作行動に対応する説明文を付与する方法、付与した結果について述べる。

%In order to realize the operative action estimation based on captioning, w
We collected texts that describe operative actions for pairs of the current and target states (images) for operative action estimation based on captioning.
We extended the Home Action Genome Dataset~\cite{rai2021home} to achieve operative action captioning for developing a robot that works in human living spaces.
In this section, we describe the Home Action Genome Dataset, our data collection method using crowdsourcing, and our collection results.

\begin{figure}[t]
    \centering
    \includegraphics[width=\linewidth]{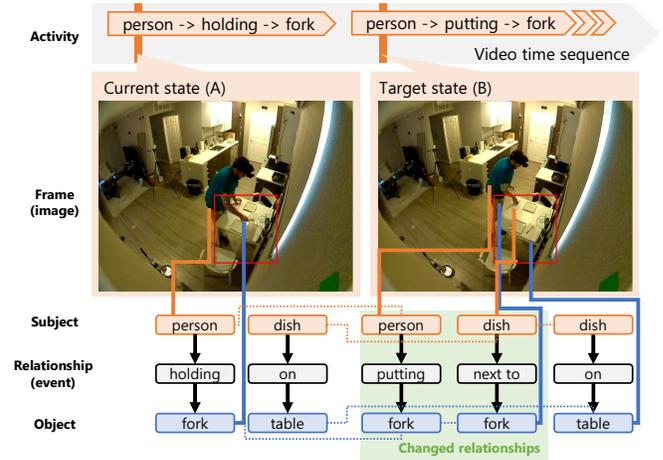}
    \caption{Annotation example of Home Action Genome Dataset}
    \label{fig:HAGdataset}
\end{figure}

\subsection{Home Action Genome Dataset}
\label{sec:data:hagenome}
% 今回ベースにしたデータセットについて、どのあたりが今回のタスクに適しているか
%Home Action Genome Dataset \cite{rai2021home} は、家庭内で人間が行う様々な行動を動画で記録し、その記録上の物体や行動、変化などをアノテーションしたデータセットである。
%動画には三人称視点の固定カメラと一人称視点の頭部カメラが存在する。
%特に三人称視点の動画には物体領域の矩形とそのラベル名、ラベルが時間変化によってどのように変化したかが記述されたアノテーションがシーングラフの形で存在する。
%シーングラフは「 subject-relationship-object 」の関係で記載され、 relationship のうち特に動作行動に関わるものは activity として定義されている。
%シーングラフ中の動作表現に関しては時間系列上のどの時点からどの時点まで動作が行われたかも付与されている。
%三人称および一人称の動画は全ての収録データに存在し時間同期が取られているため、一人称視点の動画において物体領域の矩形等はわからないものの、その中でどのような事態が起こったかがわかる。

The Home Action Genome Dataset~\cite{rai2021home} consists of videos of various human actions in the home with their annotations (Fig.~\ref{fig:HAGdataset}): subjects, objects, and relationships. 
Relationship contains relations, events, and actions.
The videos are recorded by head-mount cameras for a first-person viewpoint and fixed point cameras for a third-person viewpoint.
The third-person videos have annotations in the form of scene graphs that describe events or actions with object names, subject names, and rectangles.
Scene graphs indicate ``subject-relationship-object'' connections. 
``Activities'' are defined as events related to operative actions.
These annotations are given with a time stamp in the video's time series.
Since both the third- and first-person videos are recorded in all the sessions and are time-synchronized, these labels can be used for both videos even though we cannot use a rectangle of the object area in the first-person video.

%具体的なアノテーション例の一部を図\ref{fig:HAGdataset}に示す。
%この例では、机の上の皿の横にフォークを置くという動作に対応するようシーングラフのアノテーションが行われている。
%実際のアノテーションでは「 person-in\_front\_of-table 」などの関係も網羅的に付与されているが、この図では対象とする動作行動に関連するグラフのみに絞って記載している。

Some specific annotation examples are shown in Fig.~\ref{fig:HAGdataset}.
Here scene graphs are annotated to corresponding events and relationships to the action: putting the forks on the table by the plate.
In the actual annotation, although such relationships as ``person-in\_front\_of-table'' are comprehensively annotated, the example in Fig.~\ref{fig:HAGdataset} only shows the graphs related to the target operative action. 
%今回の動作行動推定のタスクでは、この例に記載した変化した部分、すなわち「 person-putting-fork 」と「 dish-next\_to-fork 」に着目した説明文生成を行う。

\subsection{Caption Annotation by Crowdsourcing}
\label{sec:data:crowd}
% クラウドを使ったデータ収集について

%Operative action captioning を実現するため、今回は Home Action Genome Dataset 中のある動作が行われる直前を現在の状態、動作が行われた直後の状態をターゲットの状態として利用する。
%具体的には、データセットに付与されたシーングラフのアノテーションを利用して、 activity に相当する関係が変化した時点の前後の画像フレームを抜き出して current state と target state を定義した。
%例えば図\ref{fig:HAGdataset}の例では、「 person-holding-fork 」のシーングラフが「 person-putting-fork 」に変化した前後のフレームを抜き出している。
%Home Action Genome Dataset からこうした条件に該当するフレームペアを機械的に抜き出したところ、 object として物体が関わる activity の前後でおよそ 69,000 ペアのデータ候補が抽出された。
%これらのペアに対して対象の物体に対してどのような動作行動が行われたか説明文をクラウドソーシングにより付与した。
%具体的には、2枚の画像をクラウドワーカーに提示して、画像間で対象の物体にどのような動作行動が行われたかを説明して貰った。
%クラウドワーカーに行ったインストラクションは以下の通りである。

To achieve operative action captioning, we used frames in the Home Action Genome Dataset videos.
The current state is defined as an observation just before an actual operative action, and the target state is defined as an observation just after it.
The current and target states are defined by extracting image frames before and after the point in a time sequence when the relationship changed that corresponds to the activity in the scene graph.
For example, in the example in Fig.~\ref{fig:HAGdataset}, we extracted the frames before and after the ``person-holding-fork'' scene graph changed to ``person-putting-fork.'' 
We automatically extracted approximately 69,000 pairs of candidate frames using scene graph annotation from the Home Action Genome Dataset from both first- and third-person viewpoint videos.
For each of these pairs, we added by crowdsourcing a natural language description of what kind of operative action was performed between the paired images.
We presented both images and the target object name's label extracted from the scene-graph annotation to the crowd workers and asked them to explain what kind of action was performed by the target object between the images.
The crowd workers received the following instructions.

\begin{screen}
%１枚目の画像の状態を２枚目の画像の状態にするために、作業者は何をしたでしょうか。
%物体名「〇〇」を用いて文章で説明できる場合、「説明できる」にチェックを入れ、物体名 調理台 を用いて文章で記述してください。
%説明できない場合は、「説明できない」にチェックを入れ、説明できない理由を文章で記述してください。
What did the worker in the images do to change the state in the first image to the one in the second image?
If you can explain using the ``OBJECT'', check the ``I can explain'' box and describe it using the object name.
If you cannot explain, check the ``I cannot explain'' box and describe why.
\end{screen}
%
%物体名「〇〇」では、その画像で activity が関連付けられている object の名前が入る。
%例えば図\ref{fig:overview}の例では、上の画像では「トイレ」が、下の画像では「フォーク」が入る。
OBJECT is the name of an object, which is related to the target activity.
For example, in Fig.~\ref{fig:overview}, we used ``toilet'' as the OBJECT for the upper example and ``fork'' as the OBJECT for the lower example.
Fig.~\ref{fig:overview} also indicates examples of collected captions as ``description.''

%この結果、約30,000件のペアに動作行動のアノテーションを行うことができた。
%この結果アノテーションを行うことができたペアを学習、開発、テストデータセットとして 14335：843：1686 に分割した。
%「説明できない理由」の記述からワーカーが画像を見ていないと判断できるものは非承認として再収集を行った。
%最終的に動作行動がアノテーション出来ないと判定された画像ペアには以下のようなものがあった。

Finally, we gathered 16864 pairs with operative action captions and split them into 14335/843/1686 as train/dev/test.
We checked the quality of the captions and the reasons for being unable to explain and recollected if the checking rejected the sample.
The image pairs that were finally judged as unable to be annotated included the following:

\begin{itemize}
    %\item 三人称視点画像で対象となる物体が遠すぎて判別できない
    %\item シーングラフに記述された Activity 前後で物体の位置や状態などが変化していない
    %\item 対象となる物体に人などが被っており画像からの判別が困難
    %\item 一人称視点の画像でブレが生じている
    %\item 一人称視点で視点が移動した結果、対象となる物体がいずれかの画像で映っていない
    \item The target object is too far away to be identified by the third-person viewpoint camera.
    \item The position or state of the target object did not change between the current and target images.
    \item Since the target object is obstructed by a person or other objects, distinguishing it from the image is difficult.
    \item The first-person viewpoint image is blurred.
    \item The target object is not shown in the current image, in the target image, or in either image.
\end{itemize}

\begin{figure*}[t]
    \centering
    \includegraphics[width=\linewidth]{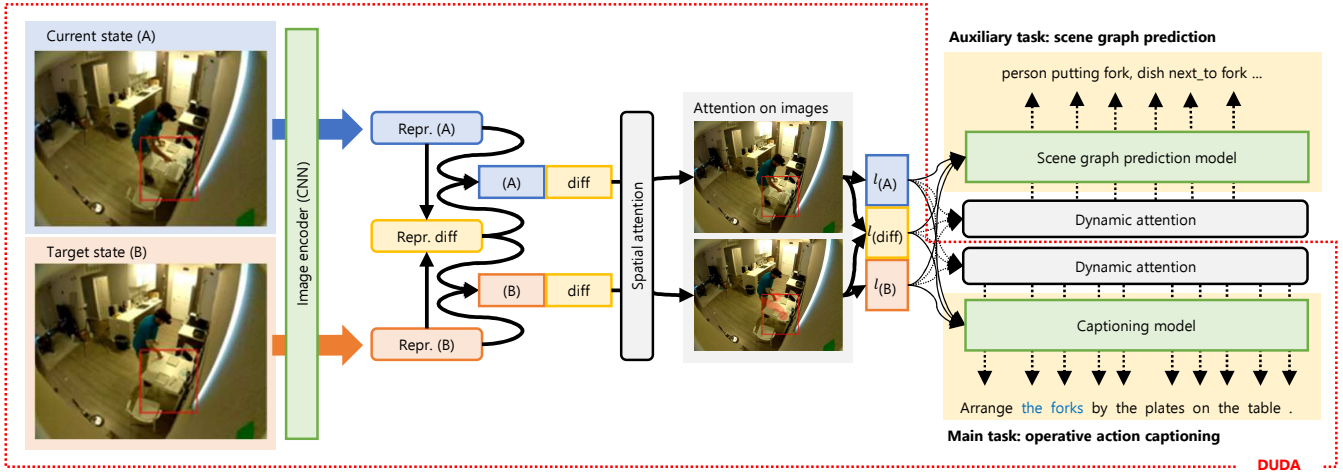}
    \caption{Dual dynamic attention (DUDA) model and its extension in the proposed system}
    \label{fig:DUDA}
\end{figure*}

\section{Operative Action Estimation Based on Natural Language Generation Model}
\label{sec:method}
% モデルの概要

%収集したデータを用いて、物体を対象とした動作行動の前後の画像から行われた（行うべき）行動を説明文として推定するモデルを構築した。
%ベースラインモデルとしては Dual Dynamic Attention (DUDA) Model~\cite{park2019robust}を用い、ここに補助タスクとしてシーングラフ推定を付与することで生成文の精度向上を試みた。
%本節では DUDA モデルの概要、補助タスクとして用いたシーングラフ推定、学習の設定などについて説明する。
%モデル概要全体を図\ref{fig:DUDA}に示す。

Using the collected data, we constructed a model to estimate the operative actions performed (or to be performed) from the images of the current and target states.
We used the dual dynamic attention (DUDA) model as our baseline scheme and improved it by adding a scene-graph prediction module as an auxiliary task.
In this section, we describe the outline of the DUDA model, the scene-graph prediction used as the auxiliary task, and the training setup.
The overall model overview is shown in Fig.~\ref{fig:DUDA}.

\subsection{Dual Dynamic Attention (DUDA) Model}
\label{sec:method:duda}
% DUDAモデルを使ったキャプショニング

%DUDAモデル\cite{park2019robust}は2枚の画像の変化に着目し、その変化内容を説明するために提案されたモデルである。
%DUDAは画像の差分検出機構を持つが、この差分検出で用いられる画像の変化部分は支援動作によって変化する物体に相当し、タスクへの相性が良いことが期待されるためこのモデルを用いる。
%モデルではまず差分検出を行うため、支援動作前と支援動作後の画像を ResNet~\cite{he2016deep} を用いた Image Encoder によって特徴ベクトルに変換する。
%Image Encoder によって抽出された各画像特徴を$X_{\sf (A)}$、$X_{\sf (B)}$として、$X_{\sf diff} = X_{\sf (A)} - X_{\sf (B)}$
%により差分ベクトルを得る。
%
The DUDA model~\cite{park2019robust}, which focuses on the change between two images, was proposed to explain the change itself.
Thus, it has an image-difference detection mechanism.
We use this model because the part of the image change extracted by difference detection corresponds to an object that changes due to the operative action, which is our task's main focus.
In the model, images before and after the operative action are converted into feature vectors by an image encoder using ResNet~\cite{he2016deep} to perform difference detection.
The extracted feature matrices of images $X_{\sf (A)}$ and $X_{\sf (B)}$ calculate the difference matrix as $X_{\sf diff} = X_{\sf (A)} - X_{\sf (B)}$.
%
%この差分ベクトルを各画像特徴に結合した$X'_{\sf (A)}$、$X'_{\sf (B)}$から差分を表す Spatial Attention $a_{\sf (A)}$、$a_{\sf (B)}$を計算する\cite{mascharka2018transparency}。
%この注視重みを各画像特徴$X_{\sf (A)}$、$X_{\sf (B)}$と差分ベクトル$X_{\sf diff}$の各要素に対して掛けることで、変化に対して重みが付けられた画像特徴である$l_{\sf (A)}$、$l_{\sf (B)}$、$l_{\sf diff}$を得る。
%この変化に対して重みが付いた画像特徴を入力として Dynamic Attention を用いた画像キャプショニングを行うことで、変化した状態に着目した画像生成が行われることを期待する。
%この際、キャプショニング結果に対する正解とのソフトマックスクロスエントロピー誤差$L_{cap}$をネットワーク学習時に用いる。
%DUDAモデルを単純に今回行うタスクに適用した場合、図\ref{fig:DUDA}のうち補助タスクである「シーングラフの推定」が存在しないモデルとなる。
The difference matrix is concatenated with the original matrices as $X'_{\sf (A)}$ and $X'_{\sf (B)}$ to compute spatial attentions $a_{\sf (A)}$ and $a_{\sf (B)}$~\cite{mascharka2018transparency}.
The elemental unit multiplication between the attention weights and feature matrices $X_{\sf (A)}$, $X_{\sf (B)}$, and $X_{\sf diff}$ are used as resultant feature vectors $l_{\sf (A)}$, $l_{\sf (B)}$, and $l_{\sf diff}$, which focus on the changes between images.
The change is also related to the operated action. 
Weighted feature vectors $l_{\sf (A)}$, $l_{\sf (B)}$, and $l_{\sf diff}$ are fed to the captioning network that has dynamic attention to generate captions.
Here our target caption indicates the conducted operative action between two images.
Softmax cross-entropy loss $L_{cap}$ for the reference caption is used as the loss function of network training.
The original DUDA model does not contain the ``auxiliary task'' part in Fig.~\ref{fig:DUDA}.

\subsection{Auxiliary Task: Scene-Graph Prediction}
\label{sec:method:auxiliary}
% 補助タスクとして用いたシーングラフ認識
%前後の画像間で行われた動作は説明文として与えられているが、説明文自体は様々な表現の揺れを含む。
%また、含まれる物体名や動作名が明示的に与えられないような説明文も存在する。
%これに対して、画像間の動作に対するアノテーションであるシーングラフの情報を補助的に用いることで、行動クラスを適切に反映した説明文を出力できることが期待できる。

The actions performed between the current and target images are given as captions, which contain various expressions.
Some captions do not explicitly contain the names of the target actions or the target objects.
The auxiliary tasks that predict the scene graphs force the network to maintain the information of the operative actions to improve the caption quality.

%このアイディアを実現するため、ベースラインであるDUDAモデルでシーングラフの推定を行う補助モデルを構築する。
%具体的には図\ref{fig:DUDA}に示す通り、変化に対して重みが付けられた画像特徴$l_{\sf (A)}$、$l_{\sf (B)}$、$l_{\sf diff}$を入力として、シーングラフを構成する三つ組を推定するモデルを考える。
%画像中から得られるシーングラフは多数存在する場合があるが、今回はネットワークに「subject-relationship-object」を繰り返し推定する系列予測モデルを構築した。
%この補助タスクは系列を行うため、本タスクと同様にソフトマックスクロスエントロピー誤差を用いる。
%この誤差を$L_{sgr}$とする。

We accomplished this idea by adding a model for the auxiliary tasks that predicts scene graphs with the baseline DUDA model.
As shown in Fig.~\ref{fig:DUDA}, we constructed a model to predict the triples in scene graphs given the same image features to the captioning model: $l_{\sf (A)}$, $l_{\sf (B)}$, and $l_{\sf diff}$.
Although many scene graphs are obtained from the images, we built a sequential prediction model that repeatedly outputs ``subject-relationship-object'' by the network.
Since this auxiliary task performs sequential prediction, we defined the network that uses the softmax cross-entropy loss as in the main task.
Let $L_{sgr}$ denote the loss function of the auxiliary task.

\subsection{Loss Function and Experimental Setup}
\label{sec:method:loss}
% 今回試す損失関数、学習の詳細な設定（学習率とか）
%まず、今回は変化を捉えるためのベースラインとしてシーングラフを行わないDUDAモデルを用いる。
%DUDAモデルは著者らの実装\footnote{\url{https://github.com/Seth-Park/RobustChangeCaptioning}}を用い、学習率の初期値は0.01、学習20エポックごとに学習率を0.1倍した。%（{\bf baseline}）。
%この損失関数は
%\begin{align}
%L_{\theta} = L_{cap} + \lambda_{L_{1,cap}} L_1 - \lambda_{ent} L_{ent,cap}, \label{Eqn:DUDA}
%\end{align}
%である。
%ここで、$L_{1,cap}$および$L_{ent,cap}$はL1正則化およびエントロピー正則化に基づく正則化項であり、$\lambda_{L_{1,cap}}$と$\lambda_{ent,cap}$はそれぞれの正則化に対する重みのハイパーパラメータである。
%このパラメータについてはDUDAで与えられているパラメータを用いた。
%この設定を{\bf ベースライン}とする。

We used the DUDA model without a scene graph as the baseline to capture the operative actions performed between both images.
We used the original implementation of the DUDA model\footnote{\url{https://github.com/Seth-Park/RobustChangeCaptioning}}.
The initial learning rate was 0.01, which was multiplied by 0.1 every 20 epochs.
The loss function is defined as,
\begin{align}
L_{\theta} = L_{cap} + \lambda_{L_{1,cap}} L_1 - \lambda_{ent} L_{ent,cap}. \label{Eqn:DUDA}
\end{align}
Here $L_{1,cap}$ and $L_{ent,cap}$ are regularizations based on L1 and entropy.
$\lambda_{L_{1,cap}}$ and $\lambda_{ent,cap}$ are the hyperparameters of the weights for each regularization.
We use the same hyperparameters from the original DUDA implementation and denote this setting as {\sf Baseline}.

%補助タスクとしてシーングラフを行う場合、ベースラインモデルであるDUDAの式(\ref{Eqn:DUDA})を以下のように拡張した。
When we trained the proposed model that has an auxiliary task to predict scene graphs, we extended the original DUDA implementation (Eq.~(\ref{Eqn:DUDA})):
\begin{align}
    L(\theta) =& \alpha (L_{cap} + \lambda_{L_{1}} L_{1,cap} - \lambda_{ent} L_{ent})\nonumber\\
    &+ (1 - \alpha) (L_{sgr} + \lambda_{L_{1,sgr}} - \lambda_{ent} L_{ent,sgr}).
\end{align}
%ここで、$\alpha$はキャプショニングとシーングラフの各タスクの損失重みである。
%その他のハイパーパラメータ$\lambda_{L_{1}}$と$\lambda_{ent}$はベースラインであるDUDAと同じものを用いた。
Here $\alpha$ is a weight that integrates the weights for operative action captioning and scene-graph prediction.
We used the same hyperparameters for the baseline for $\lambda_{L_{1}}$ and $\lambda_{ent}$.

%このとき、いくつか損失の合成法を試行した。
%1つ目は、本タスクと補助タスクの重みを固定して等価に扱う場合で、検証データで検証した上で重みとして$\alpha=0.9$とした。
%この設定を{\bf 線形補完(0.9)}とする。
%2つ目は、本タスクと補助タスクを重みをつけて交互に学習する場合である。
%この時、10エポックごとに重み$\alpha_1$と$\alpha_2$を入れ替えて学習に用いる。
%$\alpha_1$と$\alpha_2$の組み合わせは、$[0.0, 1.0], [0.1, 0.9]$の2パターンを試行した。
%この設定を{\bf 交互学習(1.0)}および{\bf 交互学習(0.9)}とする。
%なお、交互学習の1単位の中では先にシーングラフのタスクを行い、続けてキャプションの生成タスクを行った。
We implemented two integration methods for the main and auxiliary tasks.
The first is a linear interpolation that uses a fixed $\alpha$.
We set $\alpha=0.9$ based on a trial on the development dataset.
We call this setting {\sf Linear Int. (0.9)}.
Another method is a fluctuating update, which alternatively uses two $\alpha$s every ten epochs.
We tried two $\alpha$s patterns, $[0.0, 1.0]$ and $[0.1, 0.9]$, and called these settings {\sf Alternative (1.0)} and {\sf Alternative (0.9)}.
In the alternative method, we trained the scene-graph prediction task and then the captioning task.

%また、するべきシーングラフの候補には支援動作前の状態{\sf (A)}に含まれるシーングラフと支援動作後の状態{\sf (B)}に含まれるシーングラフが存在し、何を用いれば動作行動推定の精度が向上するかは明らかでない。
%そこで今回は、和集合となるシーングラフを全て推定する場合（{\bf 全体}）と、シーングラフの差分を推定する場合（{\bf 差分}）を試行した。
%これは、前者が画像全体の特徴を用いること、後者が対象となる変化、つまり動作行動にフォーカスした特徴を用いることに相当する。
In addition, we utilized two scene graph sets to investigate the best usage.
Since there are several scene graphs in the current state {\sf (A)} and the target state {\sf (B)}, we determined two setups: {\sf all} and {\sf diff}.
The {\sf all} method predicts any scene graphs contained in both {\sf (A)} and {\sf (B)}.
The {\sf diff} method predicts only the discrepancies between scene graphs {\sf (A)} and {\sf (B)}.
The former corresponds to using features from both images, and the latter corresponds to using the changes in the images.

\begin{table*}[t]
    \caption{Automatic evaluation results}
    \centering
    \begin{tabular}{l|ll|c|c|c|c|c|c} \hline
        \multirow{2}{*}{Model} &
        \multirow{2}{*}{Condition} &
        & \multicolumn{4}{c|}{BLEU} & \multirow{2}{*}{ROUGE-L} & \multirow{2}{*}{CIDEr} \\
        \cline{4-7}
        & & & 1 & 2 & 3 & 4 & & \\
        \hline \hline
        {\sf Baseline} & - &  & 0.389 & 0.238 & 0.151 & 0.0998 & 0.375 & 0.871\\
        \hline
        \multirow{6}{*}{\sf +Scene Graph} 
        & \multirow{2}{*}{\sf Alternative (1.0)} & {\sf all} & 0.350 & 0.194 & 0.113 & 0.0686 & 0.330 & 0567 \\
        \cline{3-9}
        & & {\sf diff} & 0.359 & 0.205 & 0.126 & 0.0815 & 0.339 & 0.649 \\
        \cline{2-9}
        & \multirow{2}{*}{\sf Alternative (0.9)} & {\sf all} & 0.396 & 0.244 & 0.156 & 0.105 & 0.383 & 0.921 \\
        \cline{3-9}
        & & {\sf diff} & 0.392 & 0.245 & 0.158 & 0.107 & 0.389 & 0.913 \\
        \cline{2-9}

        & \multirow{2}{*}{\sf Linear Int. (0.9)} & {\sf all} & {\bf 0.405} & {\bf 0.260} & {\bf 0.167} & {\bf 0.114} & {\bf 0.392} & {\bf 1.001} \\
        \cline{3-9}
        & & {\sf diff} & 0.396 & 0.246 & 0.160 & 0.109 & 0.387 & 0.971 \\
        \hline
        \end{tabular}
    \label{tab:autoeval_overall}
\end{table*}

\begin{table*}[t]
    \caption{Content word accuracy}
    \centering
    \begin{tabular}{l|ll|c|c|c|c|c|c|c|c|c} \hline
        \multirow{2}{*}{Model} &
        \multirow{2}{*}{Condition} &
        & \multicolumn{3}{c|}{Noun} & \multicolumn{3}{c|}{Verb} & \multicolumn{3}{c}{Verb-independent} \\
        \cline{4-12}
        & & & P & R & F & P & R & F & P & R & F\\
        \hline \hline
        {\sf Baseline} & - & & 0.378 & 0.385 & 0.381 & 0.156 & 0.169 & 0.162 & 0.122 & 0.131 & 0.126 \\ 
        \hline
        \multirow{4}{*}{\sf +Scene Graph} 
        & \multirow{2}{*}{\sf Alternative (0.9)} & {\sf all} & 0.393 & 0.406 & 0.399 & 0.168 & 0.186 & 0.177 & 0.139 & 0.153 & 0.146\\
        \cline{3-12}
        & & {\sf diff} & 0.387 & 0.406 & 0.396 & 0.164 & 0.173 & 0.168 & 0.129 & 0.138 & 0.133\\
        \cline{2-12}
        & \multirow{2}{*}{\sf Linear Int. (0.9)} & {\sf all} & {\bf 0.398} & {\bf 0.414} & {\bf 0.406} & {\bf 0.176} & {\bf 0.190} & {\bf 0.183} & {\bf 0.144} & {\bf 0.158} & {\bf 0.151}\\
        \cline{3-12}
        & & {\sf diff} & 0.387 & 0.413 & 0.400 & 0.165 & 0.180 & 0.172 & 0.140 & 0.153 & 0.146\\
        \hline
        \end{tabular}
    \label{tab:autoeval_term}
\end{table*}

\begin{table}[t]
    \caption{Human evaluation results}
    \centering
    \begin{tabular}{l|ll|c|c} \hline
        Model & Condition  & &
        Natur. & Infor. \\
        \hline \hline
        {\sf Baseline} & - & & 3.89 & 3.05 \\ 
        \hline
        \multirow{2}{*}{\sf +Scene Graph} 
        & \multirow{2}{*}{\sf Linear Int. (0.9)} & {\sf all} & 4.42 & 3.45 \\
        \cline{3-5}
        & & {\sf diff} & {\bf 4.53} & {\bf 3.48}\\
        \hline
        \end{tabular}
    \label{tab:humaneval}
\end{table}

\section{Experiments}
\label{sec:exp}
% 実験の概要、何を明らかにするのか

%実験設定で設定した各モデルを評価するため、正解データとの比較に基づく自動評価と、生成結果を人手で評価した人手評価を実施した。
%以下では評価項目、評価結果について述べる。
To evaluate each model described in the experimental setup, we conducted an automatic evaluation based on a comparison with the reference and a human evaluation in which the generated results were evaluated manually.
The evaluation criteria and results are described below.

\subsection{Evaluation Criteria}
\label{sec:exp:criteria}
% 評価手法

%まず、自動評価ではテストデータに付与されている正解との比較を行う。
%評価では n-gram 一致率に基づく BLEU-1\_4\cite{papineni2002bleu}と、最大一致長に基づくROUGE-L\cite{lin2004rouge}、一致率に単語重要度に基づく重みを加えたCIDEr\cite{vedantam2015cider}を用いる。
The automatic evaluation criteria are based on a systematic comparison with the annotated caption reference of the test set.
We used BLEU-1\_4~\cite{papineni2002bleu}, which is based on the n-gram match rate, ROUGE-L~\cite{lin2004rouge}, which is based on the maximum match length, and CIDEr~\cite{vedantam2015cider}, which is based on weighted-term matching.

%今回は動作行動推定のキャプショニングにシーングラフ推定の情報を加えるため、シーングラフに記載されている名詞や動詞などの情報がより適切に出力されていることが期待される。
%そこで、正解として付与されているキャプションのうち「名詞」「動詞」「動詞-自立」に相当する語をMeCab\cite{}によって抜き出し、これらの単語に対する適合率(P)、再現率(R)、調和平均(F)を計測する。
We expect to add the information from the scene-graph prediction to the captioning results of the motion behavior estimation; thus, we focused on nouns and verbs.
We calculated precision (P), recall (R), and harmonic mean (F) of ``nouns,'' ``verbs,'' and ``independent verbs'' extracted from the references. 

%また、自動評価には限界があるため、人手による評価も行う。
%人手評価においては、評価者に支援動作前後の状態画像（画像(A)および(B)）と生成されたキャプションを与え、キャプションの流暢性、内容の適切性について5段階で評価して貰った。
%人手評価は文の評価について訓練を受けた1名のアノテータが行った。
%人手評価はコストが掛かるため、{\bf ベースライン}と、自動評価のスコアが最も良かった{線形補完}の{\bf 全体}および{\bf 差分}の3システムに対して、テストデータからランダムに抽出した200ペアを評価した。
We also performed human evaluations because the correlations are limited between the automatic evaluation criteria and the human evaluation results.
In the human evaluation, one human evaluator was given state images (images (A) and (B)) and generated captions that described their operative actions.
The evaluator rated the caption's naturalness and informativeness~\cite{wen2015semantically} on a five-point scale.
Another annotator checked the first evaluator's result to improve the consistency.
This process was done blindly; the evaluators did not know the method names.
Because human evaluation is expensive, we evaluated 200 pairs randomly from the test set for the three systems with the best scores in the automatic evaluation: {\sf baseline}, {\sf all}, and {\sf diff} for {\sf linear completion}.

\subsection{Automatic Evaluation Results}
\label{sec:exp:result:auto}
%自動評価の結果を表\ref{tab:autoeval_overall}と表\ref{tab:autoeval_term}に示す。
%表から、基本的にはシーングラフ推定を補助タスクとして用いる方がベースラインと比較して各スコアが向上することがわかる。
%ただ、学習においては本タスクと補助タスクを完全に切り替えた交互学習よりも、相互の学習を行う際にやや重みを加えておく方が良いことが示唆された。
%また、単純な線形補完により補助タスクの影響を限定的にとどめることで、キャプションの生成をよくできることが示唆された。
Tables~\ref{tab:autoeval_overall} and \ref{tab:autoeval_term} show the automatic evaluation results.
Both tables show a primary trend, where using scene-graph predictions as auxiliary tasks improved each score compared to the baseline.
The result suggests that linear interpolation with a small weight on the auxiliary task outperforms alternative training.

%表\ref{tab:autoeval_term}の結果から、シーングラフ推定を補助タスクとして用いることで、名詞、動詞などの内容語がより出現するようになったことがわかる。
%これは、シーングラフの内容である「subject-relationship-object」の各事物が、補助タスクによってより重要視されるようになった結果であると予想される。
Based on Table~\ref{tab:autoeval_term}, our proposed method successfully generated content words, including nouns and verbs, than the baseline.
This is because the scene-graph contents represented by ``subject-relationship-object'' forced the network to keep these contents.

\begin{figure}[t]
    \centering
    \includegraphics[width=\linewidth]{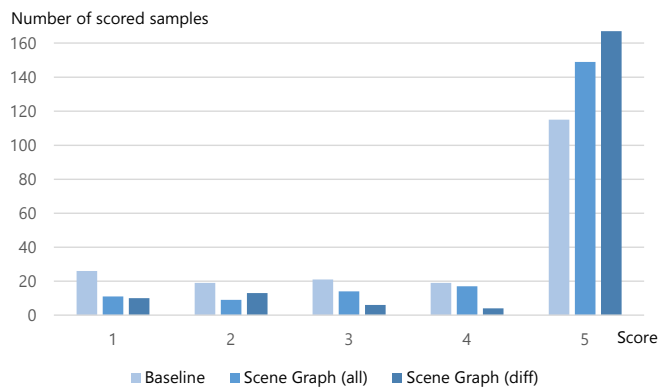}
    \caption{Distribution of naturalness score of each method}
    \label{fig:fluency}
\end{figure}

\begin{figure}[t]
    \centering
    \includegraphics[width=\linewidth]{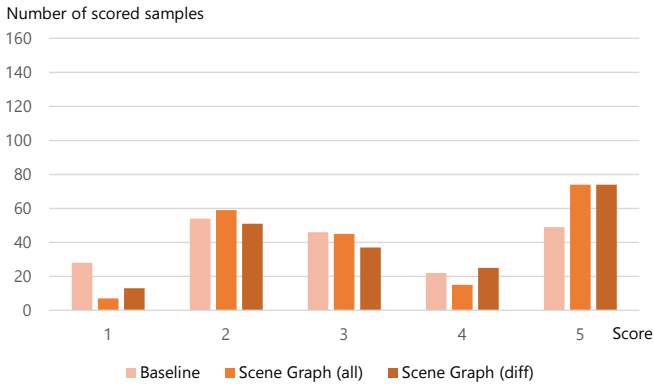}
    \caption{Distribution of informativeness score of each method}
    \label{fig:contents}
\end{figure}

\begin{figure}[t]
    \centering
    \includegraphics[width=\linewidth]{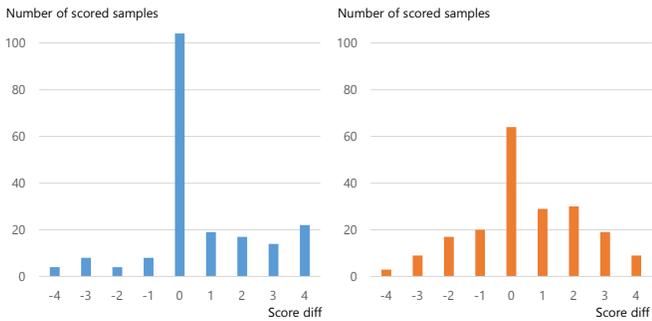}
    \caption{Distribution of naturalness (left) and informativeness (right) scores between {\sf Scene-Graph (diff)} and {\sf Baseline}}
    \label{fig:diff}
\end{figure}

\begin{figure}[t]
    \centering
    \includegraphics[width=\linewidth]{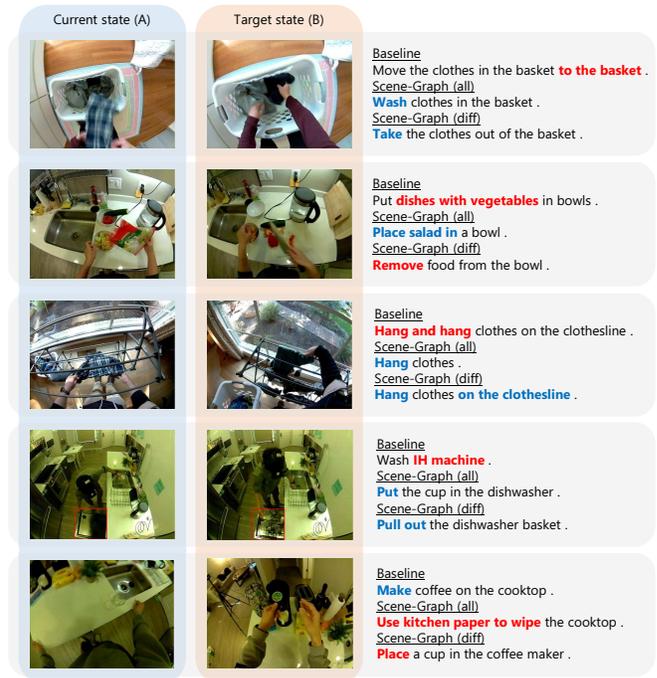}
    \caption{Case study indicating generated samples}
    \label{fig:example}
\end{figure}

\subsection{Human Evaluation Results}
\label{sec:exp:result:human}

%次に、人手評価の結果を表\ref{tab:humaneval}に示す。
%この結果から、シーングラフ推定を補助タスクに用いることが内容に関するスコア向上に寄与するだけでなく、流暢性にも大きく寄与することが示された。
%流暢性に寄与した理由として、動作行動の説明に必要不可欠な内容がシーングラフにおける「subject-relationship-object」の中身に含まれ、これらの情報が含まれない文が生成されにくくなったことが考えられる。
%また、事例分析でもある通り、キャプショニングの典型的な問題である repetition の抑圧にも効果があった可能性がある。
Table~\ref{tab:humaneval} shows the human evaluation results, which indicate that using scene-graph prediction as the auxiliary task improved the informativeness and contributed significantly to the naturalness of either our proposed method ({\sf all} or {\sf diff}).
This is probably because the auxiliary task suppressed the over-generation of content words, which are not contained in the scene graphs.
It might also suppress repetition, which is a typical problem of sentence generation, as shown in the case study analysis.

Figures~\ref{fig:fluency} and \ref{fig:contents} show the score distribution of each criterion.
The naturalness was improved by focusing on the differences in scene graphs.
Few differences exist between {\sf all} and {\sf diff} in informativeness, although they outperformed the {\sf Baseline}.

%図はベースラインと提案法の事例ごとの差の分布を示している。
%4は提案法が5点でベースラインが1点の場合、-4は提案法が1点でベースラインが5点の場合である。
%全体的な傾向としては提案法によって改善されたケースが多いが、一部提案法によってInformativenessが低下した場合もある。
%Naturalnessは多くの場合で提案法によって向上した。
Fig.~\ref{fig:diff} shows the distribution of the differences between the baseline and proposed methods ({\sf Scene graph (diff)}) for each case.
4 denotes where the proposed method scored 5 points and the baseline scored 1, and -4 denotes where the proposed method scored 1 point and the baseline scored 5.
The overall trend is that the proposed method improved the informativeness in many cases, although sometimes the proposed method decreased it.
Naturalness was improved by the proposed method in many cases.

\subsection{Case Study}
\label{sec:exp:example}

%今回の動作行動推定モデルに入力した動作前後の状態画像と、そこから生成されたキャプションの例を図\ref{fig:example}に示す。
%1つ目の例では、ベースラインでは「バスケットの中の服をバスケットに移す」という意味のない動作を説明しているが、シーングラフを補助情報に用いる提案法ではいずれも動作内容に相当するものが生成できている。
%シーングラフの全体を見るものでは「洗濯している」という動作を、差分を見るものでは「取り出している」という動作を生成している。
%いずれも誤りではないが、対象行動が一部に含まれる全体の動作を説明した前者に対し、差分に着目した後者の方がより洗濯における個別の動作を説明している。

As a case study, Fig.~\ref{fig:example} shows four examples of the current state ({\sf A}), the target state ({\sf B}), and the captions from three methods: {\sf Baseline}, {\sf Scene Graph (all)}, and {\sf Scene Graph (diff)}.
In the first example, the baseline describes a meaningless and redundant action of ``move the clothes in the basket to the basket,'' although the proposed methods with scene-graph prediction as an auxiliary task generated actual actions.
{\sf All}, which looks at the entire scene graph, generated action ``wash,'' and {\sf diff}, which looks at the differences in the scene graphs, generated a more detailed action: ``take out.''
Although both are correct, the former described the overall action in which the target action was included; the latter which focused on the differences described a specific action included in the idea of washing.

%2つ目の例では、ベースラインは「ボウルに野菜の入った食器を入れる」という実際ではあまり起こらない行動を生成してしまっているが、提案法ではその点が改善されている。
%しかし、差分に注目したケースでは「出す」という実際の行動とは逆向けの内容を生成してしまっている。
%意味的に逆の動作は埋め込み空間上でも近傍に配置されやすく、こうしたケースに対する対処は今後検討する必要がある。
In the second example, the baseline generated an action that rarely occurs in reality, ``put dishes with vegetables in bowls,'' which was improved in the proposed methods.
However, when we focused on the differences ({\sf diff}), the proposed method generated an opposite action (``remove'') instead of the actual action (``place'').
Since opposite actions tend to be placed near each other in the embedding space, we must consider how to deal with such cases in future work.

%3つ目の例では、ベースラインではrepititionが起こってしまっているが、補助情報を使うことによって問題が抑圧されている。
%また差分に着目することで、対象動作の object である「物干し」が文中に生成されている。
In the third example, the baseline caused a repetition problem, which was suppressed by the proposed methods.
Focusing on the difference might generate an object name ``clothesline.''

%4つ目の例では、「洗う」という動作はいずれも正しいが、全体を見る場合と差分を見る場合では説明された行動の粒度が異なっている。
%特に差分は「食洗器でコップを洗う」という正解の行動の一部として含まれる「食洗器のかごを引き出す」という動作を説明している。
%ロボットでの動作生成のような応用を考えた場合、こうした動作の粒度は適切に制御できるようになることが望ましい。
In the fourth example, both proposed methods successfully generated actions corresponding to using a dishwasher; however, they explained with different granularity.
In particular, {\sf diff} explained the ``pull out the dishwasher basket'' action, which is included as part of the overall ``put the cup in the dishwasher'' action.
When we apply these methods to a human-assisting scenario at home, we must discuss the controllability of the granularity of the captions.
%considering applications such as motion generation in robots, it is desirable to be able to control the granularity of such actions appropriately.

In the fifth example, only the baseline successfully captioned the action of ``make coffee.''
This is probably because the proposed models tried to use objects in scene graph and failed.

%\section{Related Work}
%\label{sec:relatedworks}

\section{Conclusion}
\label{sec:conc}

%本研究では、ロボットなどの人間を支援するシステムが状況と理想状態から行うべき動作行動を推定するため、この動作行動推定をキャプショニングによって言語化する枠組みを構築した。
%具体的には、支援動作前後の画像に対してその間で行われた支援動作に相当するキャプションをクラウドソーシングで付与し、このキャプションを生成するモデルの構築を行った。
%この際、動作行動に関連した物体や動作そのものに着目するため、各状態におけるシーングラフを補助的に推定して用いるモデルを提案した。
%自動・人手評価を用いた実験の結果、提案法は補助情報を推定することでより適切な動作行動推定を行うことができるようになったことが示された。
%今後は今回構築したモデルから実際にロボットの支援動作行動を生成する枠組みを構築する。
%In this study, in order to estimate the operative action of human-assistive systems such as robots, w
We constructed a framework to verbalize the required operative actions given both the current and target (ideal) states by captioning networks to estimate the operative action of such human-assistive systems as robots.
We constructed a dataset by crowdsourcing that consists of triplets of a current state, a target state, and a caption that describes the operative action to change the current state to the target state.
We proposed a captioning model that uses scene-graph prediction as an auxiliary task by focusing on object names and the relationships represented in scene graphs.
We investigated the effect of our proposed method through both automatic and human evaluations, especially on human evaluation results on naturalness and informativeness.
Our future work will include the proposed system in our human-assisting robot at home~\cite{yuguchi2022butsukusa}.

\bibliography{survey}

\begin{thebibliography}{10}
\providecommand{\url}[1]{#1}
\csname url@rmstyle\endcsname
\providecommand{\newblock}{\relax}
\providecommand{\bibinfo}[2]{#2}
\providecommand\BIBentrySTDinterwordspacing{\spaceskip=0pt\relax}
\providecommand\BIBentryALTinterwordstretchfactor{4}
\providecommand\BIBentryALTinterwordspacing{\spaceskip=\fontdimen2\font plus
\BIBentryALTinterwordstretchfactor\fontdimen3\font minus
  \fontdimen4\font\relax}
\providecommand\BIBforeignlanguage[2]{{%
\expandafter\ifx\csname l@#1\endcsname\relax
\typeout{** WARNING: IEEEtran.bst: No hyphenation pattern has been}%
\typeout{** loaded for the language `#1'. Using the pattern for}%
\typeout{** the default language instead.}%
\else
\language=\csname l@#1\endcsname
\fi
#2}}

\bibitem{vinyals2016show}
O.~Vinyals, A.~Toshev, S.~Bengio, and D.~Erhan, ``Show and tell: Lessons
  learned from the 2015 mscoco image captioning challenge,'' \emph{IEEE
  Transactions on Pattern Analysis and Machine Intelligence}, vol.~39, no.~4,
  pp. 652--663, 2016.

\bibitem{you2016image}
Q.~You, H.~Jin, Z.~Wang, C.~Fang, and J.~Luo, ``Image captioning with semantic
  attention,'' in \emph{Proceedings of the IEEE/CVF Conference on Computer
  Vision and Pattern Recognition (CVPR)}, 2016, pp. 4651--4659.

\bibitem{dou2018data2text}
L.~Dou, G.~Qin, J.~Wang, J.-G. Yao, and C.-Y. Lin, ``Data2text studio:
  Automated text generation from structured data,'' in \emph{Proceedings of the
  2018 Conference on Empirical Methods in Natural Language Processing (EMNLP):
  System Demonstrations}, 2018, pp. 13--18.

\bibitem{ishigaki2021generating}
T.~Ishigaki, G.~Topi{\'c}, Y.~Hamazono, H.~Noji, I.~Kobayashi, Y.~Miyao, and
  H.~Takamura, ``Generating racing game commentary from vision, language, and
  structured data,'' in \emph{Proceedings of the 14th International Conference
  on Natural Language Generation (INLG)}, 2021, pp. 103--113.

\bibitem{takano2015statistical}
W.~Takano and Y.~Nakamura, ``Statistical mutual conversion between whole body
  motion primitives and linguistic sentences for human motions,'' \emph{The
  International Journal of Robotics Research}, vol.~34, no.~10, pp. 1314--1328,
  2015.

\bibitem{yamada2018paired}
T.~Yamada, H.~Matsunaga, and T.~Ogata, ``Paired recurrent autoencoders for
  bidirectional translation between robot actions and linguistic
  descriptions,'' \emph{IEEE Robotics and Automation Letters}, vol.~3, no.~4,
  pp. 3441--3448, 2018.

\bibitem{yoshino2020caption}
K.~Yoshino, K.~Wakimoto, Y.~Nishimura, and S.~Nakamura, ``Caption generation of
  robot behaviors based on unsupervised learning of action segments,''
  \emph{Conversational Dialogue Systems for the Next Decade}, p. 227, 2020.

\bibitem{yuguchi2022butsukusa}
A.~Yuguchi, S.~Kawano, K.~Yoshino, C.~T. Ishi, Y.~Kawanishi, Y.~Nakamura,
  T.~Minato, Y.~Saito, and M.~Minoh, ``Butsukusa: A conversational mobile robot
  describing its own observations and internal states,'' in \emph{Proceedings
  of the 2022 ACM/IEEE International Conference on Human-Robot Interaction
  (HRI)}, 2022, pp. 1114--1118.

\bibitem{soans2020sa}
N.~Soans, E.~Asali, Y.~Hong, and P.~Doshi, ``Sa-net: Robust state-action
  recognition for learning from observations,'' in \emph{Proceedings of the
  2020 IEEE International Conference on Robotics and Automation (ICRA)}.\hskip
  1em plus 0.5em minus 0.4em\relax IEEE, 2020, pp. 2153--2159.

\bibitem{ahn2022can}
M.~Ahn, A.~Brohan, N.~Brown, Y.~Chebotar, O.~Cortes, B.~David, C.~Finn,
  K.~Gopalakrishnan, K.~Hausman, A.~Herzog, \emph{et~al.}, ``Do as i can, not
  as i say: Grounding language in robotic affordances,'' \emph{arXiv preprint
  arXiv:2204.01691}, 2022.

\bibitem{chatila2018toward}
R.~Chatila, E.~Renaudo, M.~Andries, R.-O. Chavez-Garcia, P.~Luce-Vayrac,
  R.~Gottstein, R.~Alami, A.~Clodic, S.~Devin, B.~Girard, \emph{et~al.},
  ``Toward self-aware robots,'' \emph{Frontiers in Robotics and AI}, vol.~5,
  no.~88, 2018, {DOI}:10.3389/frobt.2018.00088.

\bibitem{kim2019image}
D.-J. Kim, J.~Choi, T.-H. Oh, and I.~S. Kweon, ``Image captioning with very
  scarce supervised data: Adversarial semi-supervised learning approach,'' in
  \emph{Proceedings of the 2019 Conference on Empirical Methods in Natural
  Language Processing and the 9th International Joint Conference on Natural
  Language Processing (EMNLP-IJCNLP)}, 2019, pp. 2012--2023.

\bibitem{gan2017semantic}
Z.~Gan, C.~Gan, X.~He, Y.~Pu, K.~Tran, J.~Gao, L.~Carin, and L.~Deng,
  ``Semantic compositional networks for visual captioning,'' in
  \emph{Proceedings of the IEEE/CVF Conference on Computer Vision and Pattern
  Recognition (CVPR)}, 2017, pp. 5630--5639.

\bibitem{yao2017incorporating}
T.~Yao, Y.~Pan, Y.~Li, and T.~Mei, ``Incorporating copying mechanism in image
  captioning for learning novel objects,'' in \emph{Proceedings of the IEEE/CVF
  Conference on Computer Vision and Pattern Recognition (CVPR)}, 2017, pp.
  6580--6588.

\bibitem{li2019pointing}
Y.~Li, T.~Yao, Y.~Pan, H.~Chao, and T.~Mei, ``Pointing novel objects in image
  captioning,'' in \emph{Proceedings of the IEEE/CVF Conference on Computer
  Vision and Pattern Recognition (CVPR)}, 2019, pp. 12\,497--12\,506.

\bibitem{li2017scene}
Y.~Li, W.~Ouyang, B.~Zhou, K.~Wang, and X.~Wang, ``Scene graph generation from
  objects, phrases and region captions,'' in \emph{Proceedings of the IEEE
  International Conference on Computer Vision (ICCV)}, 2017, pp. 1261--1270.

\bibitem{tran2018closer}
D.~Tran, H.~Wang, L.~Torresani, J.~Ray, Y.~LeCun, and M.~Paluri, ``A closer
  look at spatiotemporal convolutions for action recognition,'' in
  \emph{Proceedings of the IEEE/CVF Conference on Computer Vision and Pattern
  Recognition (CVPR)}, 2018, pp. 6450--6459.

\bibitem{ji2020action}
J.~Ji, R.~Krishna, L.~Fei-Fei, and J.~C. Niebles, ``Action genome: Actions as
  compositions of spatio-temporal scene graphs,'' in \emph{Proceedings of the
  IEEE/CVF Conference on Computer Vision and Pattern Recognition (CVPR)}, 2020,
  pp. 10\,236--10\,247.

\bibitem{rai2021home}
N.~Rai, H.~Chen, J.~Ji, R.~Desai, K.~Kozuka, S.~Ishizaka, E.~Adeli, and J.~C.
  Niebles, ``Home action genome: Cooperative compositional action
  understanding,'' in \emph{Proceedings of the IEEE/CVF Conference on Computer
  Vision and Pattern Recognition (CVPR)}, 2021, pp. 11\,184--11\,193.

\bibitem{chen2020say}
S.~Chen, Q.~Jin, P.~Wang, and Q.~Wu, ``Say as you wish: Fine-grained control of
  image caption generation with abstract scene graphs,'' in \emph{Proceedings
  of the IEEE/CVF Conference on Computer Vision and Pattern Recognition
  (CVPR)}, 2020, pp. 9962--9971.

\bibitem{park2019robust}
D.~H. Park, T.~Darrell, and A.~Rohrbach, ``Robust change captioning,'' in
  \emph{Proceedings of the IEEE/CVF International Conference on Computer Vision
  (CVPR)}, 2019, pp. 4624--4633.

\bibitem{qiu20203d}
Y.~Qiu, Y.~Satoh, R.~Suzuki, K.~Iwata, and H.~Kataoka, ``3d-aware scene change
  captioning from multiview images,'' \emph{IEEE Robotics and Automation
  Letters}, vol.~5, no.~3, pp. 4743--4750, 2020.

\bibitem{kim2021fixmypose}
H.~Kim, A.~Zala, G.~Burri, and M.~Bansal, ``Fixmypose: Pose correctional
  captioning and retrieval,'' in \emph{Proceedings of the AAAI Conference on
  Artificial Intelligence}, vol.~35, no.~14, 2021, pp. 13\,161--13\,170.

\bibitem{he2016deep}
K.~He, X.~Zhang, S.~Ren, and J.~Sun, ``Deep residual learning for image
  recognition,'' in \emph{Proceedings of the IEEE/CVF Conference on Computer
  Vision and Pattern Recognition (CVPR)}, 2016, pp. 770--778.

\bibitem{mascharka2018transparency}
D.~Mascharka, P.~Tran, R.~Soklaski, and A.~Majumdar, ``Transparency by design:
  Closing the gap between performance and interpretability in visual
  reasoning,'' in \emph{Proceedings of the IEEE/CVF Conference on Computer
  Vision and Pattern Recognition (CVPR)}, 2018, pp. 4942--4950.

\bibitem{papineni2002bleu}
K.~Papineni, S.~Roukos, T.~Ward, and W.-J. Zhu, ``Bleu: a method for automatic
  evaluation of machine translation,'' in \emph{Proceedings of the 40th annual
  meeting of the Association for Computational Linguistics (ACL)}, 2002, pp.
  311--318.

\bibitem{lin2004rouge}
C.-Y. Lin, ``Rouge: A package for automatic evaluation of summaries,'' in
  \emph{Text summarization branches out}, 2004, pp. 74--81.

\bibitem{vedantam2015cider}
R.~Vedantam, C.~Lawrence~Zitnick, and D.~Parikh, ``Cider: Consensus-based image
  description evaluation,'' in \emph{Proceedings of the IEEE conference on
  computer vision and pattern recognition (CVPR)}, 2015, pp. 4566--4575.

\bibitem{wen2015semantically}
T.-H. Wen, M.~Gasic, N.~Mrk{\v{s}}i{\'c}, P.-H. Su, D.~Vandyke, and S.~Young,
  ``Semantically conditioned lstm-based natural language generation for spoken
  dialogue systems,'' in \emph{Proceedings of the 2015 Conference on Empirical
  Methods in Natural Language Processing (EMNLP)}, 2015, pp. 1711--1721.

\end{thebibliography}
\bibliographystyle{IEEEtran}

\end{document}